%% file: main.tex
\documentclass[10pt,twocolumn,letterpaper]{article}

\usepackage{cvpr}              
\usepackage{graphicx}
\usepackage{booktabs}
\usepackage{multirow} 
\input{preamble}
\definecolor{cvprblue}{rgb}{0.21,0.49,0.74}
\usepackage[pagebackref,breaklinks,colorlinks,allcolors=cvprblue]{hyperref}

\def\paperID{*****} 
\def\confName{CVPR}
\def\confYear{2026}

\title{Semantic and Visual Evidence for Efficient Long-Video Reasoning:\\ A Solution for the HD-EPIC VQA Challenge\thanks{Team Name: HIPPO team}}

\author{
Yinsong Xu \quad Wei Jing \quad Liuxin Zhang \quad Wanjun Lv \quad Hui Li\\
Lenovo, China\\
{\tt\small \{xuys10, jingwei1, zhanglx2, lvwj1, lihuid\}@lenovo.com}
}

\begin{document}
\maketitle

\input{sec/0_abstract}    
\input{sec/1_intro}
\input{sec/2_method}
\input{sec/3_experiment}

\input{sec/4_conclusion}
{
    \small
    \bibliographystyle{ieeenat_fullname}
    \bibliography{main}
}


\end{document}

%% file: sec/0_abstract.tex
\begin{abstract}
Understanding long-form egocentric videos remains challenging for multimodal large language models (MLLMs) due to limited context length and insufficient grounding of fine-grained visual details. The recently proposed HD-EPIC benchmark highlights these limitations: even strong long-context models achieve relatively low performance across diverse video question answering tasks. In this paper, we propose a unified framework that decouples long-video reasoning into two complementary forms of evidence: \emph{semantic evidence} and \emph{visual evidence}. Semantic evidence captures global procedural structure through a coarse-to-fine extraction pipeline, while object-centric visual evidence preserves fine-grained grounding through bounding boxes and visual embeddings. During inference, we formulate reasoning as a query-conditioned evidence retrieval and integration process, dynamically selecting relevant information from both sources. Our approach achieves competitive performance in the HD-EPIC-VQA Challenge across multiple task categories. More broadly, our results demonstrate that explicitly structuring, retrieving, and integrating semantic and visual evidence is critical for effective long-video understanding with MLLMs.
\end{abstract}

%% file: sec/1_intro.tex
\section{Introduction}
Understanding long-form egocentric videos~\cite{yang2025egolife,grauman2022ego4d,perrett2025hdepic} remains a central challenge for multimodal large language models (MLLMs). Real-world videos, such as cooking demonstrations, span long temporal horizons and contain complex interactions among objects, actions, and environments. Reasoning over such videos therefore requires both global procedural understanding and fine-grained visual grounding.

\begin{figure}[t]
  \centering
  \includegraphics[width=\linewidth]{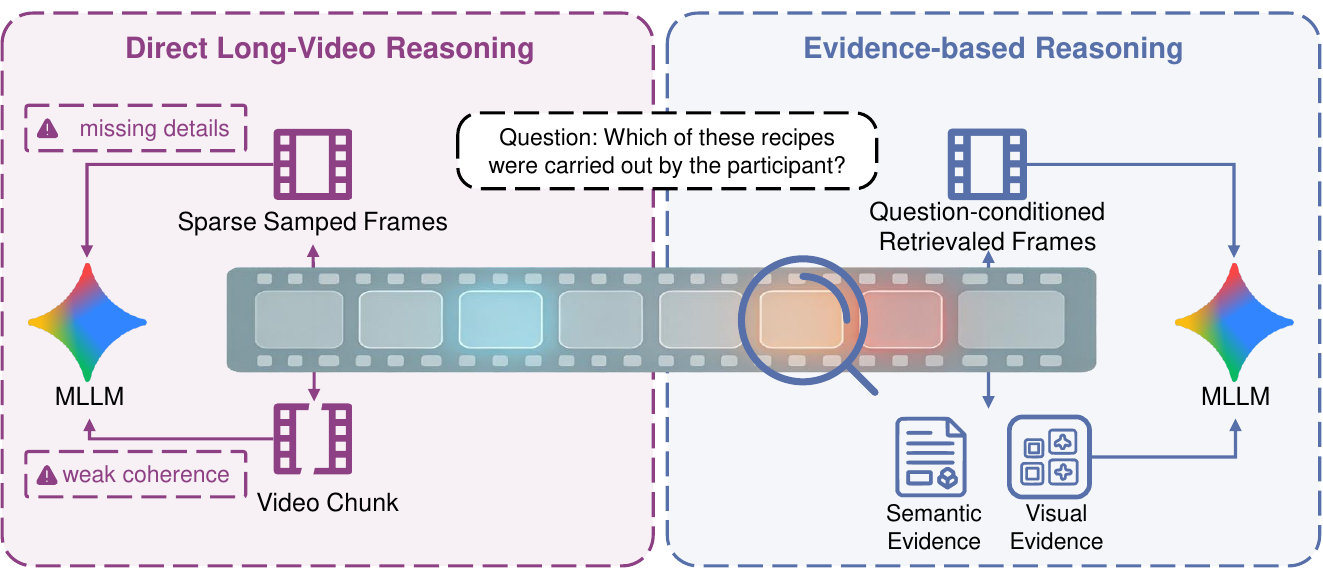}
\caption{Comparison between direct long-video reasoning and our evidence-guided reasoning framework. \textbf{Left:} Direct frame sampling from a long video either weakens global coherence or misses fine-grained details under the limited context capacity of MLLMs. \textbf{Right:} Our framework constructs reusable \emph{semantic evidence} and \emph{visual evidence} from the full video. Given a question, the framework retrieves relevant evidence and selects a compact set of task-relevant frames. The retrieved evidence and selected frames are then jointly provided to the MLLM for answer prediction.}

\label{fig:teaser}
\end{figure}

Current MLLMs struggle in this setting because long-video inputs often exceed their effective context capacity. Existing methods typically address this limitation through sparse frame sampling~\cite{zhang2024videoinstructiontuningsynthetic,Maaz2023VideoChatGPT} or chunk-based processing~\cite{rege2026agentic}. As illustrated in \cref{fig:teaser}, these strategies fail in complementary ways: sparse sampling discards fine-grained visual evidence, whereas chunk-based processing weakens global temporal coherence and introduces substantial computational overhead.

\begin{figure*}[t]
  \centering
  \includegraphics[width=\linewidth]{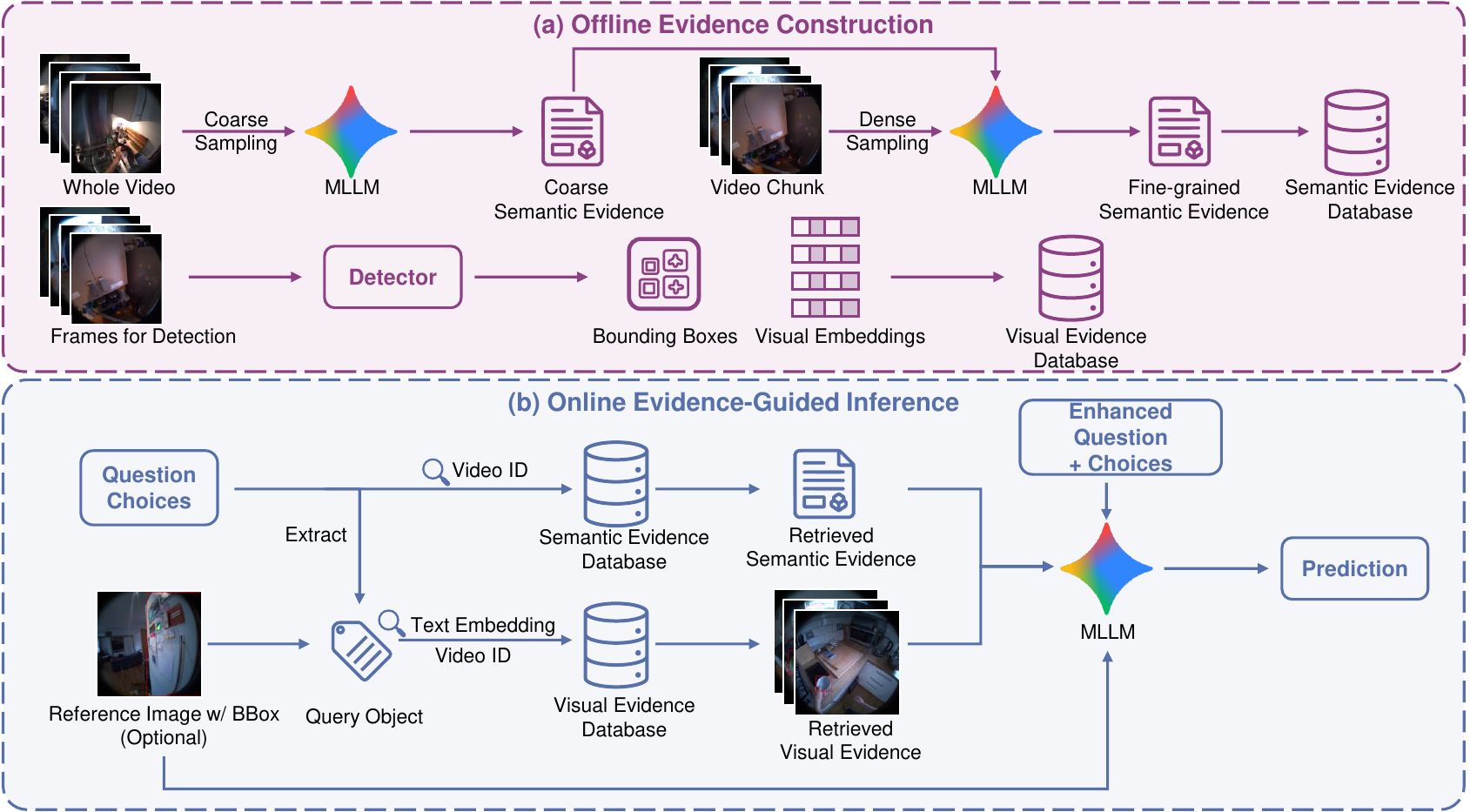}
  \caption{Overview of the proposed two-stage framework. (a) Offline construction builds reusable semantic evidence through coarse-to-fine MLLM-based summarization and constructs an object-centric visual evidence database from bounding-box proposals and visual embeddings. (b) Online evidence-guided inference retrieves semantic and visual evidence conditioned on the question, choices, video ID, and optional reference image with bounding box. The retrieved evidence is used to select task-relevant frames, which are then provided to the MLLM together with the retrieved evidence and enhanced prompts for prediction.}
  \label{fig:framework}
\end{figure*}

To address these limitations, we decouple long-video reasoning into two stages: offline query-agnostic evidence construction and online evidence-guided inference, as shown in \cref{fig:teaser}. In the first stage, we construct two complementary forms of reusable evidence from the full video. \emph{Semantic evidence} is extracted by an MLLM through a coarse-to-fine pipeline and represents global procedural context, visible operations, and tool interactions as structured text. \emph{Visual evidence} is constructed using an object detector and preserves object-level bounding boxes and visual embeddings for spatial grounding. In the second stage, given an input question, a query-conditioned evidence retrieval mechanism identifies relevant semantic and visual evidence and uses the retrieved evidence to select a compact set of task-relevant frames. The retrieved evidence and selected frames are then jointly provided to the MLLM for answer prediction. Compared with direct sampling over the entire video, this design substantially reduces the number of input frames and mitigates interference from redundant or irrelevant visual content.

Our approach was developed for the HD-EPIC VQA Challenge~\cite{perrett2025hdepic}, a large-scale benchmark comprising 41 hours of egocentric kitchen videos and 26K questions. Our method achieves competitive performance in the challenge across multiple task categories. These results suggest that explicitly structuring, retrieving, and integrating semantic and visual evidence is a promising direction for scaling MLLMs to long-form, real-world video understanding.

%% file: sec/2_method.tex
\section{Method}
\label{sec:method}
\subsection{Overview}
As shown in \cref{fig:framework}, our approach consists of two stages. In the offline query-agnostic construction stage, we build two complementary evidence databases to reduce the computational cost and context-length burden of reasoning over long raw videos. The semantic evidence database stores structured textual information, including recipe-level and activity-level descriptions (\cref{sec:memory}). The visual evidence database stores object-centric information, including bounding boxes and visual embeddings (\cref{sec:visual}). In the online evidence-guided inference stage, the question, choices, and optional reference image with bounding box are used to retrieve evidence from the two databases. The retrieved evidence is further used to select a compact set of task-relevant frames. The selected frames, retrieved evidence, and enhanced questions and choices are then jointly fed into the MLLM to produce the final prediction (\cref{sec:inference}).

\subsection{Semantic Evidence}
\label{sec:memory}
Directly processing entire long-form videos with MLLMs is often infeasible due to limited context length. In practice, this constraint forces the model to rely on sparsely sampled frames, which leads to incomplete temporal coverage and the loss of critical information. A straightforward workaround is to split the video into chunks and process them independently. However, chunk-level processing tends to overemphasize local details while missing global contextual information. As shown in \cref{fig:framework}(a), we therefore build a coarse-to-fine semantic evidence extraction pipeline.

The coarse-grained phase extracts high-level information, including candidate recipe names, coarse ingredients, activity stages, and major temporal transitions. Specifically, we sample the entire video at a low temporal resolution into a fixed number of frames. The MLLM then generates a global summary, which provides guidance for the fine-grained phase.

The fine-grained phase performs denser temporal observation. We split the video into fixed-length, non-overlapping chunks and sample frames more frequently. For each chunk, the MLLM takes the sampled frames and the global summary as input, and predicts structured information, including visible operations, involved ingredients, tool interactions, ingredient addition events, and candidate step boundaries.

The resulting semantic evidence is stored in the semantic evidence database and indexed by video ID. This design is important for challenge settings in which tasks require access to long-term information. In such cases, semantic evidence serves as reusable external context, avoiding repeated full-video analysis during inference.

\subsection{Visual Evidence}
\label{sec:visual}
Semantic evidence captures procedural and activity-level information but may omit fine-grained visual details, making it insufficient for questions involving specific objects, such as object localization or fixture interaction. To address this limitation, we introduce an object-centric visual evidence database, as shown in \cref{fig:framework}(a). Specifically, for each video frame, we extract bounding-box proposals and their corresponding visual embeddings, and store them in the visual evidence database together with the video ID and timestamp.

\subsection{Evidence-Guided Inference}
\label{sec:inference}
In the evidence-guided inference stage, as shown in \cref{fig:framework}(b), input construction is task-dependent but follows a common paradigm: semantic evidence provides procedural context, visual evidence provides object-centric observations, and the question is enhanced to better leverage the reasoning capability of the MLLM. The retrieved evidence further guides the selection of a compact set of task-relevant frames, which are provided to the MLLM together with the retrieved evidence.

For memory-heavy tasks, the MLLM primarily consumes semantic evidence retrieved by video ID. For example, ingredient ordering can be solved using the chronological ingredient additions stored in the semantic evidence database. For object-related tasks, such as object localization, we first extract a query object from the question. When the question provides a reference image with a bounding box, we ask an MLLM to identify the query object. The object text is then encoded by a text encoder. We compute the similarity between the text embedding and all bounding-box proposals in the input video. When a frame contains multiple proposals, we take the maximum similarity score and retain the timestamp of the matched item if the score exceeds the threshold $\tau$. Formally, for a query object with embedding $e$, the retrieved evidence $\mathcal{E}$ is:
\begin{align}
    \mathcal{E} = \{t \mid \max_{b \in B_t }{ \text{cos}(b, e)} > \tau\}, 
\end{align}
where $t$ denotes the timestamp, $B_t$ denotes the set of detected bounding-box proposals at timestamp $t$, and each $b \in B_t$ denotes the visual embedding of a detected proposal. $\text{cos}(\cdot,\cdot)$ denotes the similarity function between the proposal embedding and the query-object embedding. To improve robustness, the model may output multiple query terms, such as synonyms or more general object names. We merge matches from multiple query terms, deduplicate timestamps, and optionally retain the associated bounding boxes as localized evidence. For hybrid tasks that involve multiple videos, we first use semantic evidence to select likely videos and narrow the search space, while retrieved or sampled frames are used to verify fine-grained evidence.

%% file: sec/3_experiment.tex
\section{Experiments}
\subsection{Implementation Details}
In the offline evidence construction stage, we use Gemini~3.1~Pro to construct semantic evidence. For coarse semantic evidence, we process 600-second video chunks sampled at 0.1 FPS to capture the overall scene, main tasks, and key temporal transitions. For fine-grained semantic evidence, we process 60-second video chunks sampled at 1 FPS to record detailed action steps, object state changes, and precise temporal anchors.

For visual evidence, we use WeDetect-Large-Uni~\cite{fu2025wedetect} as the object detector with a detection threshold of 0.3, and extract visual evidence at 1 FPS. During online evidence-guided inference, we use Gemini~3.1~Pro to recognize the query object from the reference image with bounding box when available. We then use the wedetect-large-text-encoder~\cite{fu2025wedetect} to extract the query-object text embedding, and set the similarity threshold to $\tau=0.2$ for matching against visual evidence.

Depending on the question type, we provide the MLLM with retrieved frames, retrieved semantic evidence, or both. This task-dependent input construction supplies the model with the most relevant evidence while avoiding unnecessary processing overhead.

\input{table/result}
\subsection{Main Results}
\cref{tab:detail} compares our method with several representative baselines, including general-purpose MLLMs (VideoLLaMA2~\cite{damonlpsg2024videollama2}, LLaVA-Video~\cite{zhang2025llavavideovideoinstructiontuning}, and Gemini Pro) and the specialized long-video reasoning system DeepFrames~\cite{yang2026optimizingmultimodalllmsegocentric}. Following the official evaluation protocol, the overall score is computed as the average of category-level accuracies, assigning equal weight to each category. Our method achieves an overall accuracy of \textbf{65.8\%}, outperforming all baselines.

Compared with prior approaches, our method consistently improves performance across nearly all tasks, with particularly large margins on complex temporal reasoning problems. For example, we achieve substantial gains on \textit{Multi-Step Localization} (98.0\% vs.\ 88.0\% of Gemini Pro and 70.0\% of DeepFrames), \textit{Prep Localization} (78.0\% vs.\ 50.0\%), and \textit{Nutrition Change} (82.0\% vs.\ $\leq$ 26.0\% for all baselines). These tasks require modeling long-range temporal dependencies and tracking state transitions, suggesting that semantic evidence effectively captures structured procedural information for long-video understanding.

Our method also shows strong improvements on tasks requiring fine-grained spatial grounding. For instance, we outperform DeepFrames by 14\% on \textit{Object Location} (64.0\% vs.\ 49.8\%) and by +16.3 points on \textit{Fixture Location} (48.2\% vs.\ 34.2\%). Similarly, in \textit{Object Movement Itinerary}, our method achieves 49.0\% accuracy, compared to at most 18.0\% for prior MLLMs. These results highlight the importance of visual evidence, which provides object-centric representations for precise spatial reasoning.

Finally, even on challenging tasks where all methods perform relatively poorly, such as \textit{Exact Ingredient Recognition} (54.0\% vs.\ $\leq$ 38.0\%), our approach still achieves the best performance. While part of the performance gain can be attributed to the strength of the underlying foundation models, the consistent improvements across diverse task types suggest that the proposed evidence-guided framework plays a key role in enabling effective long-form video reasoning.

%% file: table/result.tex
\begin{table*}[t]
\centering
\caption{Per task accuracy (\%).
  Baseline numbers are from~\cite{yang2026optimizingmultimodalllmsegocentric}
  and~\cite{perrett2025hdepic}.
  \textbf{Bold} denotes the best result per row.}
\label{tab:detail}
\resizebox{\textwidth}{!}{%
\begin{tabular}{llccccc}
\toprule
\textbf{Category} & \textbf{Task}
  & \textbf{VideoLLaMA2}
  & \textbf{LLaVA-Video}
  & \textbf{Gemini Pro}
  & \textbf{DeepFrames}
  & \textbf{Ours} \\
\midrule
\multirow{8}{*}{\textbf{Recipe}}
& Following Activity Recognition  & 64.0 & 62.0 & 54.0 & 70.0 & \textbf{84.0} \\
& Multi-Recipe Recognition        & 52.0 & 68.0 & 76.0 & 72.0 & \textbf{84.0} \\
& Multi-Step Localization         & 18.0 & 44.0 & 88.0 & 70.0 & \textbf{98.0} \\
& Prep Localization               & 13.0 & 21.0 & 35.0 & 50.0 & \textbf{78.0} \\
& Recipe Recognition              & 22.0 & 28.0 & 42.0 & 34.0 & \textbf{86.0} \\
& Rough Step Localization         & 21.0 & 24.0 & 74.0 & 73.0 & \textbf{90.0} \\
& Step Localization               & 38.0 & 20.0 & 70.0 & 68.0 & \textbf{88.0} \\
& Step Recognition                & 13.0 & 23.0 & 45.0 & 81.0 & \textbf{90.0} \\
\midrule
\multirow{6}{*}{\textbf{Ingredient}}
& Ingredient Retrieval            & 19.0 & 22.0 & 49.0 & 76.0 & \textbf{88.0} \\
& Ingredient Weight               & 30.0 & 36.0 & 46.0 & 32.0 & \textbf{92.0} \\
& Ingredients Order               & 20.0 & 38.0 & 56.0 & 36.0 & \textbf{76.0} \\
& Ingredient Adding Localization  & 27.0 & 41.0 & 62.0 & 48.0 & \textbf{85.0} \\
& Ingredient Recognition          & 26.0 & 36.0 & 36.0 & 30.0 & \textbf{76.0} \\
& Exact Ingredient Recognition    & 32.0 & 28.0 & 28.0 & 38.0 & \textbf{54.0} \\
\midrule
\multirow{3}{*}{\textbf{Nutrition}}
& Image Nutrition Estimation      & 24.0 & 28.0 & 26.0 & 25.0 & \textbf{50.0} \\
& Nutrition Change                & 20.0 & 26.0 & 16.0 & 26.0 & \textbf{82.0} \\
& Video Nutrition Estimation      & 54.0 & 62.0 & 62.0 & 60.0 & \textbf{86.0} \\
\midrule
\multirow{4}{*}{\textbf{Action}}
& How Recognition                 & 25.2 & 41.4 & 35.6 & 36.6 & \textbf{67.6} \\
& Why Recognition                 & 32.2 & 51.2 & 43.2 & 51.0 & \textbf{75.4} \\
& Action Localization             & 20.7 & 20.9 & 30.3 & 24.9 & \textbf{41.7} \\
& Action Recognition              & 30.9 & 58.6 & 49.3 & 55.6 & \textbf{76.8} \\
\midrule
\multirow{4}{*}{\textbf{3D Perception}}
& Fixture Interaction Counting    & 17.7 & 16.3 & 35.3 & 29.0 & \textbf{46.0} \\
& Fixture Location                & 18.8 & 21.8 & 20.8 & 34.2 & \textbf{48.2} \\
& Object Location                 & 31.0 & 30.6 & 32.4 & 49.8 & \textbf{64.0} \\
& Object Contents Retrieval       & 35.5 & 40.5 & 41.5 & 50.5 & \textbf{58.5} \\
\midrule
\multirow{3}{*}{\textbf{Object Motion}}
& Object Movement Itinerary       & 11.0 &  9.8 & 18.0 & 14.2 & \textbf{49.0} \\
& Object Movement Counting        & 44.0 & 20.0 & 13.0 & 44.5 & \textbf{51.0} \\
& Stationary Object Localization  & 30.5 & 27.0 & 31.5 & 31.0 & \textbf{58.0} \\
\midrule
\multirow{2}{*}{\textbf{Gaze}}
& Gaze Estimation                 & 30.0 & 47.5 & 36.5 & 51.4 & \textbf{61.3} \\
& Interaction Anticipation        & 12.4 & 11.1 & 20.8 & 14.5 & \textbf{38.9} \\
\bottomrule
\end{tabular}}
\end{table*}

%% file: sec/4_conclusion.tex
\section{Conclusion}
We presented an evidence-guided framework for long-form egocentric video question answering on the HD-EPIC-VQA benchmark. By decoupling long-video reasoning into reusable semantic evidence and object-centric visual evidence, our method enables MLLMs to combine global procedural context with fine-grained visual grounding while avoiding exhaustive video processing at inference time. Our framework achieves competitive performance in the HD-EPIC-VQA Challenge, demonstrating the effectiveness of structured evidence construction and query-conditioned retrieval for long-form video understanding.